%% file: main.tex
\documentclass[10pt,twocolumn,letterpaper]{article}

\usepackage[pagenumbers]{cvpr} 

\input{preamble}

\usepackage{caption}
\usepackage{enumitem}
\usepackage{diagbox}
\usepackage{booktabs}
\usepackage{multirow}
\usepackage{color,colortbl}
\usepackage{subcaption}
\usepackage{rotating}
\definecolor{cvprblue}{rgb}{0.21,0.49,0.74}
\usepackage[pagebackref,breaklinks,colorlinks,citecolor=cvprblue]{hyperref}

\DeclareMathOperator*{\agga}{agg_1}
\DeclareMathOperator*{\aggb}{agg_2}

\definecolor{Gray}{gray}{0.9}
\definecolor{White}{rgb}{1,1,1}

\def\paperID{9581} 
\def\confName{CVPR}
\def\confYear{2024}

\title{YolOOD: Utilizing Object Detection Concepts \\for Multi-Label Out-of-Distribution Detection}

\author{
Alon Zolfi\textsuperscript{1},
Guy Amit\textsuperscript{1},
Amit Baras\textsuperscript{1},
Satoru Koda\textsuperscript{2},
Ikuya Morikawa\textsuperscript{2},
Yuval Elovici\textsuperscript{1},
Asaf Shabtai\textsuperscript{1}\\
\\
\textsuperscript{1}Ben-Gurion University of the Negev, Israel\\
{\tt\small \{zolfi,guy5,barasa\}@post.bgu.ac.il,\{elovici,shabtaia\}@bgu.ac.il}\\
\textsuperscript{2}Fujitsu Limited, Japan\\
{\tt\small \{koda.satoru,morikawa.ikuya\}@fujitsu.com}
}

\begin{document}
\maketitle
\input{sections/00_abstract}
\input{sections/01_introduction}
\input{sections/02_background}
\input{sections/03_method}

\input{sections/04_evaluation}
\input{sections/05_related_work}
\input{sections/06_conclusion}
{
    \small
    \bibliographystyle{ieeenat_fullname}
    \bibliography{main}
}


\end{document}

%% file: preamble.tex
\usepackage[dvipsnames]{xcolor}


%% file: sections/00_abstract.tex
\begin{abstract}
\vspace{-0.15cm}
Out-of-distribution (OOD) detection has attracted a large amount of attention from the machine learning research community in recent years due to its importance in deployed systems.
Most of the previous studies focused on the detection of OOD samples in the multi-class classification task.
However, OOD detection in the multi-label classification task, a more common real-world use case, remains an underexplored domain.
In this research, we propose YolOOD -- a method that utilizes concepts from the object detection domain to perform OOD detection in the multi-label classification task.
Object detection models have an inherent ability to distinguish between objects of interest (in-distribution) and irrelevant objects (e.g., OOD objects) in images that contain multiple objects belonging to different class categories. 
These abilities allow us to convert a regular object detection model into an image classifier with inherent OOD detection capabilities with just minor changes.
We compare our approach to state-of-the-art OOD detection methods and demonstrate YolOOD's ability to outperform these methods on a comprehensive suite of in-distribution and OOD benchmark datasets.

\end{abstract}

%% file: sections/01_introduction.tex
\vspace{-0.2cm}
\section{\label{sec:intro}Introduction}

Machine learning and particularly deep learning-based networks have become a state-of-the-art solution for computer vision tasks, such as image classification~\cite{krizhevsky2012imagenet,he2015delving}, object detection~\cite{redmon2018yolov3,ren2015faster}, and image segmentation~\cite{chen2014semantic,chen2019hybrid}.
However, it has been shown that these models can produce overconfident predictions on samples that are not within the distribution they were trained on, \ie, OOD samples~\cite{nguyen2015deep}.

In the last few years, many solutions have been proposed to address this problem, most of which focus on the separation of in-distribution and OOD data~\cite{liang2018enhancing,lee2018simple,huang2021importance,amit2021food}.
However, these studies only proposed solutions for the multi-class classification task, in which an input is associated with a single class category.
The problem of OOD detection in the \emph{multi-label} classification domain has been overlooked and remains underexplored.
Despite its importance, only two studies specifically addressed this problem~\cite{hendrycks2019scaling,wang2021can}.

Multi-label image classification and object detection are two closely related tasks.
The former involves assigning multiple labels to an image, while the latter goes a step further by not only recognizing the objects present in an image but also localizing them with bounding boxes.
Therefore, object detectors have the inherent ability to distinguish between objects of interest and irrelevant objects~\cite{ren2015faster,redmon2018yolov3,liu2016ssd}.
This ability, along with object detection similarity to multi-label classification, can be leveraged to create an OOD detection mechanism for the multi-label setting.

\begin{figure}[t!] 
    \centering
    \scalebox{0.85}{
    \begin{minipage}{0.99\linewidth}
        \centering
        \begin{sideways}
            \begin{minipage}[c]{0.23\textwidth}
                \hspace{11pt}
                \scriptsize
                In-distribution
            \end{minipage}
        \end{sideways}
        \begin{subfigure}{0.29\linewidth}
            \centering
            \includegraphics[width=0.97\linewidth]{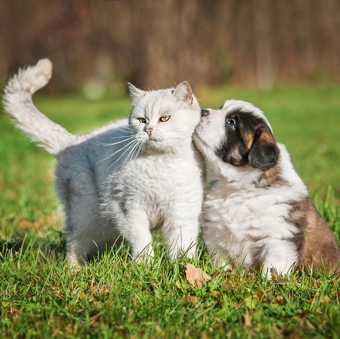}
        \end{subfigure}
        \begin{subfigure}{0.29\linewidth}
            \centering
            \includegraphics[width=0.97\linewidth]{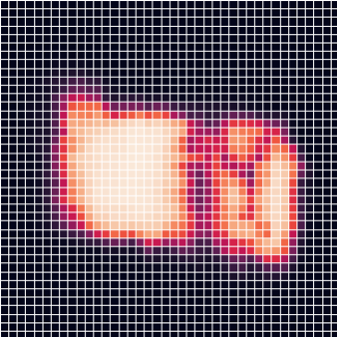}
        \end{subfigure}
        \begin{subfigure}{0.29\linewidth}
            \centering
            \includegraphics[width=0.97\linewidth]{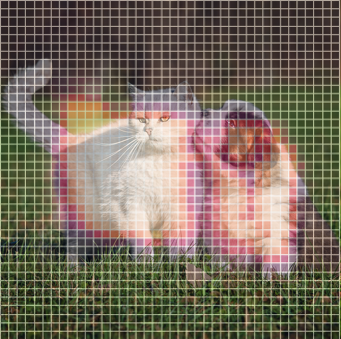}
        \end{subfigure}\\
        \vspace{2pt}
        \begin{sideways}
            \begin{minipage}[c]{0.31\textwidth}
                \hspace{16pt}
                \scriptsize
                Out-of-distribution
            \end{minipage}
        \end{sideways}
        \begin{subfigure}{0.29\linewidth}
            \centering
            \includegraphics[width=0.965\linewidth]{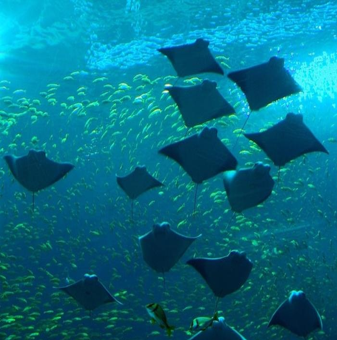}
            \caption{}
        \end{subfigure}
        \begin{subfigure}{0.29\linewidth}
            \centering
            \includegraphics[width=0.97\linewidth]{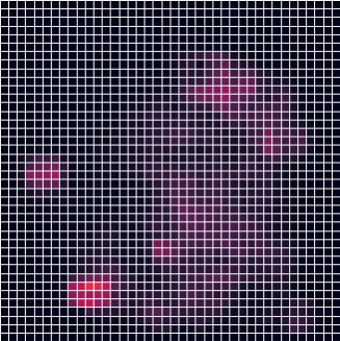}
            \caption{}
        \end{subfigure}
        \begin{subfigure}{0.29\linewidth}
            \centering
            \includegraphics[width=0.97\linewidth]{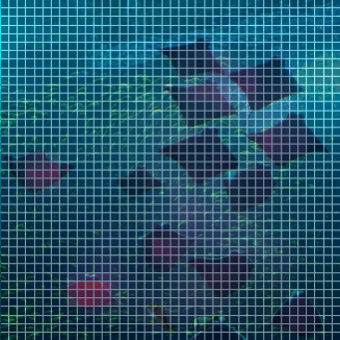}
            \caption{}
        \end{subfigure}
    \end{minipage}
    \hspace{-10pt}
    \begin{minipage}{0.03\linewidth}
        \begin{sideways}
            \includegraphics[width=19.5\linewidth]{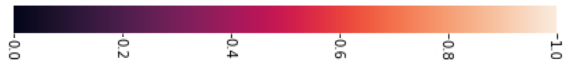}
            \hspace{-14pt}
        \end{sideways}
    \end{minipage}}
    \caption{Examples of (a) in-distribution (top) and OOD (bottom) images with (b) the corresponding YolOOD confidence heatmap, and (c) heatmap placed on-top of the image.}
    \label{fig:intro}
    \vspace{-0.2cm}
\end{figure}

In this paper, we propose \emph{YolOOD}, a multi-label classifier that utilizes the main concepts of state-of-the-art object detectors, and specifically, the YOLO object detector~\cite{redmon2018yolov3,bochkovskiy2020yolov4,yolov5}.
To convert YOLO's network into an image classifier, we simply replace the last layer of each detection head with a simplified one.
\emph{YolOOD} is based on the objectness score concept, which is commonly used in object detectors~\cite{ren2015faster,redmon2016you} to understand the relevance of the different parts of an image.
This concept is realized by training the network to predict low scores for background areas or areas that contain irrelevant objects.
In terms of OOD detection, this can be interpreted as assigning low scores to OOD data and high scores to in-distribution data, as shown in Figure~\ref{fig:intro}.

As opposed to existing OOD methods, in which the networks are actively trained on negative data from an external OOD data source~\cite{hendrycks2018deep,mohseni2020self}, YolOOD offers a major advantage, as it exploits all parts of the image to model both the original data distribution and objects that might be OOD, without the need for external data sources.
To tackle our approach's requirement of bounding box annotations, we leverage the tremendous progress large language models (LLMs) have gained lately to perform fully-automated dataset labeling.
Using Grounding DINO~\cite{liu2023grounding}, a multi-modal open-set object detection model that can detect arbitrary objects with textual inputs, we automatically generate bounding boxes for each image with merely standard image classification annotations (\ie, category names).

We perform extensive evaluations (more than 500 trained models) which demonstrate YolOOD's state-of-the-art performance and superiority over commonly used OOD detection methods on large-scale benchmark image datasets (\eg, MS-COCO~\cite{lin2014microsoft}).
In addition, we propose new in-distribution and OOD dataset benchmarks for OOD detection in the multi-label domain.
To create the in-distribution dataset, we extract a subset of 20 classes from the Objects365 dataset.
With regard to the OOD datasets, we aim to create datasets that better capture the complexity of the multi-label setting in which images may contain multiple objects belonging to different class categories. 
The new datasets consist of: 
(a) a subset of the Objects365 dataset~\cite{shao2019objects365} in which the classes are different than the classes in the in-distribution subset mentioned above, and 
(b) a subset of the NUS-WIDE dataset~\cite{chua2009nus}.
Our results show that YolOOD substantially improves the FPR95 compared to other OOD detection methods (\eg, when using the Objects365 subset as the in-distribution dataset and the NUS-WIDE subset as the OOD dataset, the FPR95 value decreases by 12.27\%).

\noindent We summarize our contributions as follows:
\begin{itemize}
    \item We propose YolOOD, a novel OOD detection technique for the multi-label image classification domain, which is powered by the YOLO object detection system and outperforms state-of-the-art techniques.
    \item We are the first to adopt the use of bounding box annotations for OOD detection - a unique technique that exploits all parts of an input image to model both the original data distribution and OOD data, without depending on external data sources.
    \item We introduce new benchmark datasets: 
    (a) a large-scale in-distribution dataset, and 
    (b) two new OOD datasets that better reflect the complexity of OOD detection in the multi-label domain (\ie, images may contain multiple objects of different class categories) and make them available to the scientific community.
\end{itemize}

%% file: sections/02_background.tex
\section{\label{sec:background}Background}

\subsection{\label{subsec:background:multi_label}Multi-Label Classification}

The multi-label classification problem is defined as follows: let $\mathcal{X}$ be the input space and $\mathcal{Y}$ be the output space of classifier ${f: \mathcal{X} \to \mathbb{R}^{\vert \mathcal{Y}\vert}}$ trained on samples drawn from distribution $\mathcal{D}(\mathcal{X},\mathcal{Y})$.
Every input $x \in \mathcal{X}$ can be associated with a subset of labels $\mathcal{Y}=\{1,2,...,N_c \}$, which is represented by a binary indicator vector $y=\{0, 1\}^{N_c}$, where $y_n=1$ if the input is associated with class $n$.

Object detection is a variant of the multi-label classification task, in which the model also determines the location of existing objects, \ie, bounding box coordinates.

\subsection{\label{subsec:background:OODD}Out-of-Distribution Detection}

Let $\mathcal{D}_{\text{in}}$ denote the marginal distribution of $\mathcal{D}$ over $X$, which represents the distribution of in-distribution data.
At inference time, the system may encounter an input drawn from a different distribution $\mathcal{D}_{\text{out}} \text{ over } \mathcal{X}$.
For OOD detection in the multi-label setting, a decision function $G$ is defined such that:
\begin{equation}
G(x;f)=
    \begin{cases}
    1 & \text{if } x \sim \mathcal{D}_{\text{in}}\\
    0 & \text{if } x \sim \mathcal{D}_{\text{out}}
    \end{cases}
\end{equation}
where $x$ is considered OOD if none of the objects present in it are in-distribution objects.

\subsection{\label{subsec:background:OD}YOLO Object Detector}

In this paper, we focus on the state-of-the-art one-stage YOLO object detector and leverage its capabilities, and our proposed approach utilizes one of its latest versions, YOLOv5~\cite{yolov5}. 

\noindent\textbf{YOLO's architecture.}
YOLO's architecture is comprised of two components: 
(a) a backbone network used to extract features from the input image, and 
(b) three detection heads which process the image's features at three different scales.
The detection heads' sizes are determined by the size of the input image and the network's \emph{stride} (downsampling factor) -- 32, 16, and 8.
This allows the network to detect objects of different sizes: the first detection head (with the largest stride) has a broader context, specializing in the detection of large objects, while the smallest one has better resolution and specializes in the detection of small objects.

The last layer of each detection head predicts a 3D tensor of size $W \times H \times (4+1+N_c)$, where $W \times H$ is the grid size ($W$ and $H$ are the width and height, respectively) and $4+1+N_c$ (which will be referred to as a \emph{candidate} in the remainder of the paper) encodes three parts: the bounding box offsets, objectness score, and class scores.

\noindent\textbf{Training YOLO.}
Every ground-truth object is associated with a single cell in each detection head.
The input image is divided into an $W \times H$ grid, and the responsible cell is determined by the grid cell that the object's center falls in.
YOLO does not assume that the class categories are mutually exclusive; therefore the class score vector is trained under the multi-label configuration (the sigmoid function is applied on each output neuron).

\noindent\textbf{Postprocessing.}
YOLO outputs a fixed amount of candidates (the amount depends on the size of the input image) which are then filtered in three sequential steps: objectness score filtering, class score filtering, and non-maximum suppression (NMS).

Additional details about YOLO's architecture, training, and inference can be found in the supplementary material.

%% file: sections/03_method.tex
\section{\label{sec:method}Method}

Object detectors provide a natural solution for OOD detection because they have an inherent ability to distinguish between objects of interest (in-distribution data) and irrelevant objects (OOD data).
Additionally, object detectors use \emph{passive negative learning} during training, utilizing the unlabeled data present in the training images (\ie, areas that are not included within labeled bounding boxes), which allows them to better generalize to the sub-task of discarding irrelevant objects.
in YOLO, this is accomplished using the objectness score.
Overall, the combination of these factors makes object detectors perfect candidates for OOD detection tasks.

In this section, we introduce \emph{YolOOD}, our novel OOD detection method for the multi-label domain, inspired by the YOLO object detector~\cite{yolov5}.
With just minor changes, we convert a YOLO network into a multi-label image classifier.

\subsection{\label{subsec:method:det_layer}YolOOD Detection Layer}

Since we propose a method for the image classification domain, our model does not have to predict bounding box coordinates.
Therefore, we replace the last layer of each detection head with a 3D tensor of size ${W_k \times H_k \times (1 + N_c)}$, where $k$ denotes the $k^{\text{th}}$ detection head ($k \in \{1,2,3\}$).
The size $W_k \times H_k$ grid remains identical to the grid in the original YOLO architecture, where each cell only predicts the objectness score and $N_c$ class scores.

Formally, let $f_{\text{YolOOD}}:\mathcal{X}\to\mathcal{C}$ be a \emph{YolOOD} image classifier that receives an input image $x \in \mathcal{X}$ and outputs a set of candidates $f_{\text{YolOOD}}(x)=\mathcal{C}$ such that $\mathcal{C}=\bigcup_{k}\mathcal{C}_k$, where $\mathcal{C}_k$ denotes the candidates of the $k^{\text{th}}$ detection head.
Each set of candidates $\mathcal{C}_k$ is comprised of ${W_k \times H_k}$ candidates (based on the downsampling factor mentioned in Section~\ref{sec:background}), and thus $\vert \mathcal{C} \vert = \sum_{k=1}^3 \vert \mathcal{C}_k \vert = \sum_{k=1}^3 W_k \cdot H_k$.

\paragraph{\label{subsec:method:obj_score}Objectness Score.}
\begin{figure}[t!]
    \centering
    \begin{subfigure}{0.4\linewidth}
        \centering
        \includegraphics[width=0.95\linewidth]{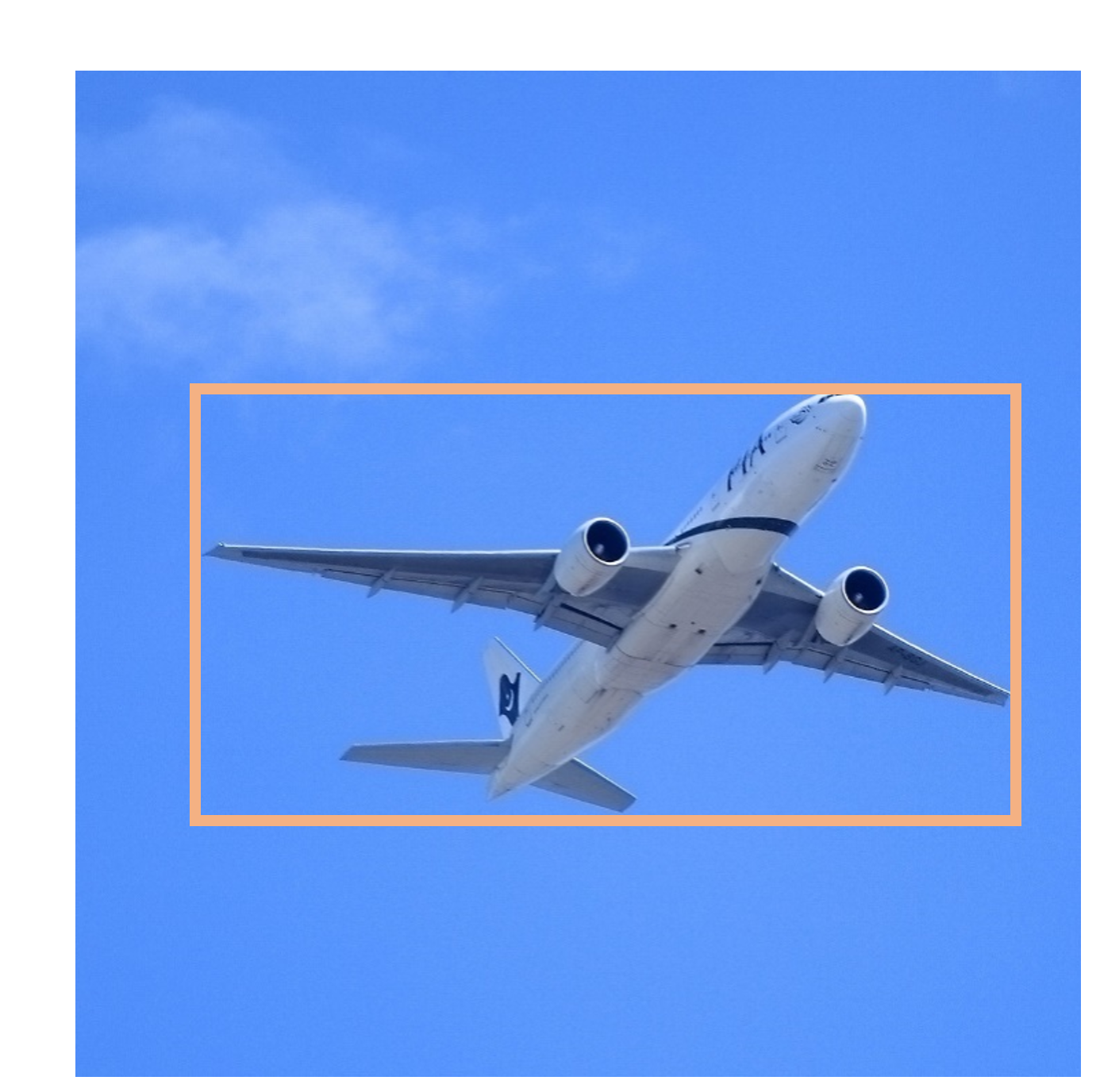}
        \caption{Clean}
    \end{subfigure}
    \begin{subfigure}{0.4\linewidth}
        \centering
        \includegraphics[width=0.95\linewidth]{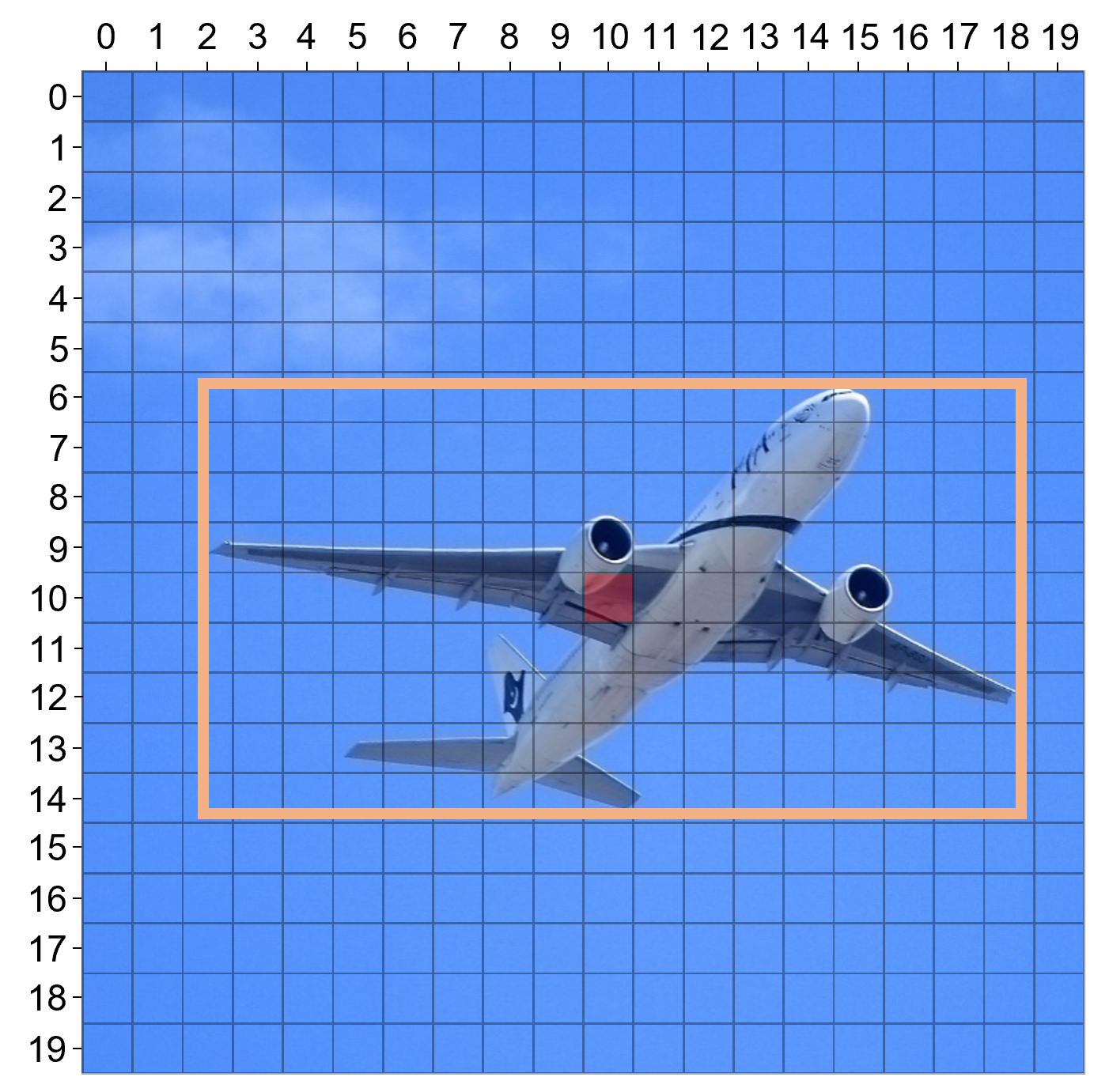}
        \caption{$p=0$}
        \label{subfig:digital_blue}
    \end{subfigure}
    \begin{subfigure}{0.4\linewidth}
        \centering
        \includegraphics[width=0.95\linewidth]{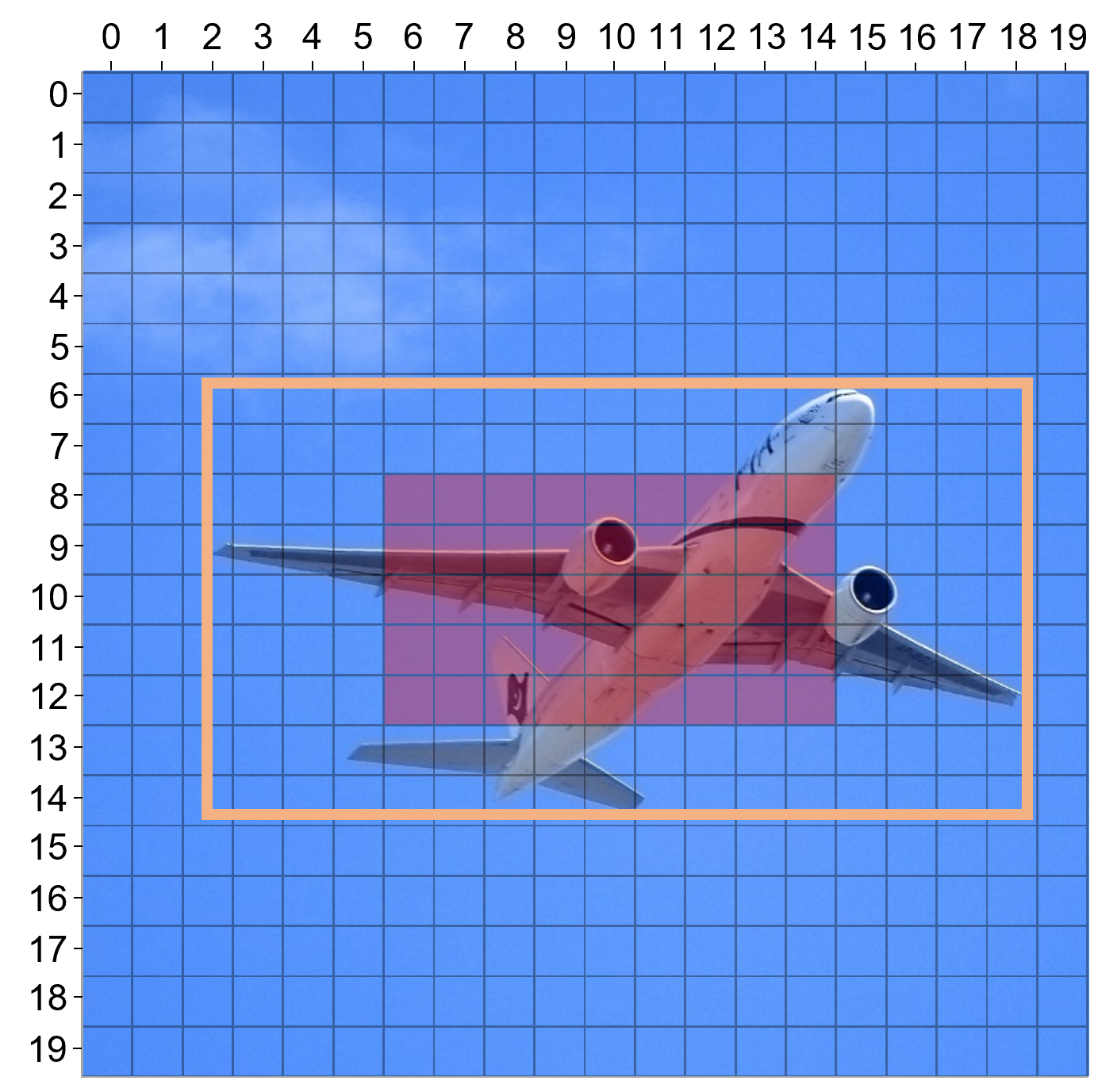}
        \caption{$p=0.5$}
    \end{subfigure}
    \begin{subfigure}{0.4\linewidth}
        \centering
        \includegraphics[width=0.95\linewidth]{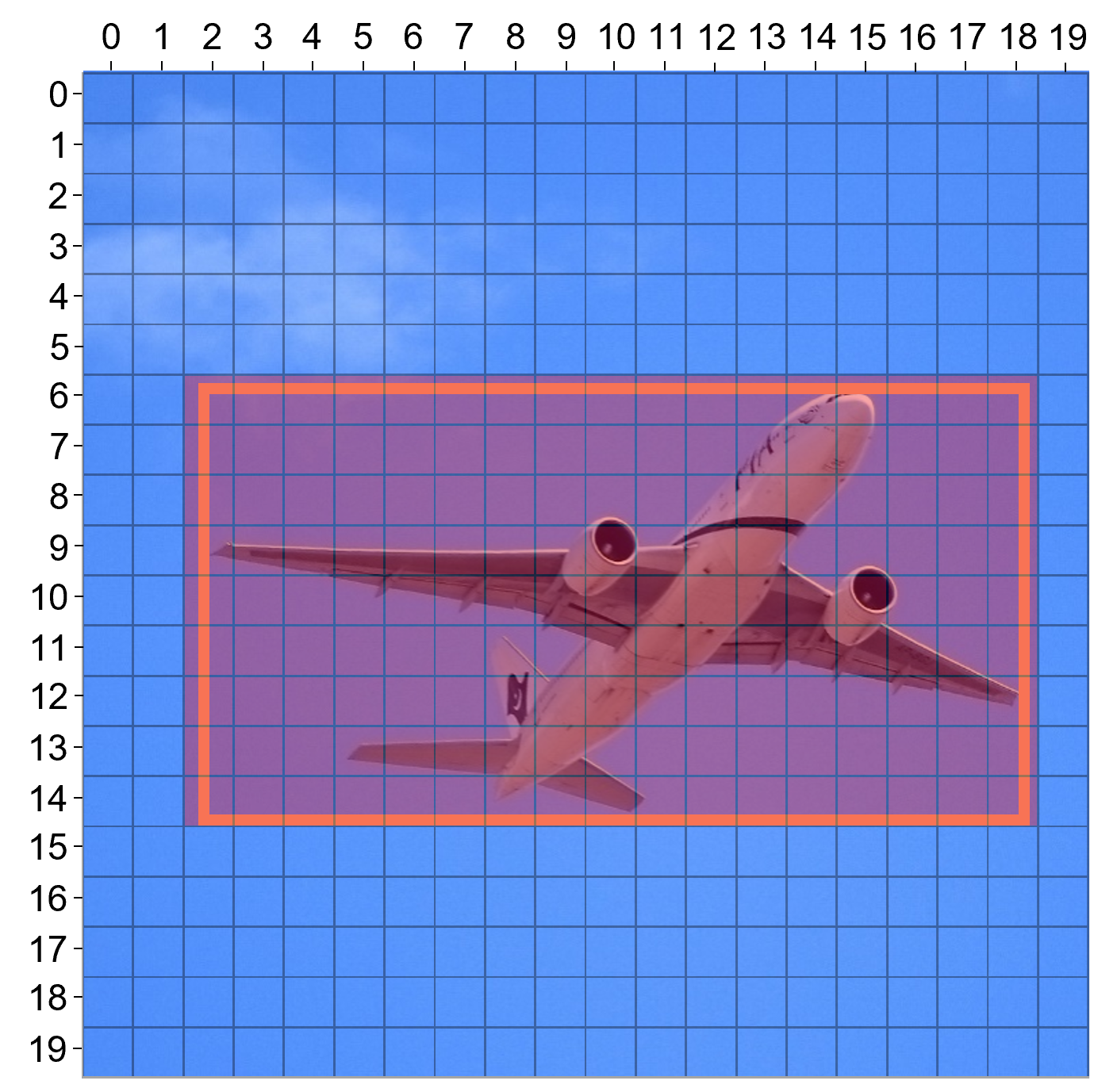}
        \caption{$p=1$}
    \end{subfigure}
    \caption{An example of an image divided into a size ${20 \times 20}$ grid with various $p$ values ($k$ is omitted for simplicity).
    The ground-truth bounding box is framed in orange.
    Grid cells highlighted in red represent the responsible cells (explained in Section~\ref{subsec:method:obj_score}).}
    \label{fig:p_obj}
\end{figure}
As explained in Section~\ref{sec:background}, in the original YOLO model, only a single cell grid is responsible for the prediction of an object.
However, in image classification, we are not limited to the case of a specific cell predicting the object's bounding box.
Therefore, we propose enlarging the ``responsibility" areas, such that multiple grid cells which contain \emph{any} part of an object are responsible for its detection, allowing the model to capture broader representations of objects.

However, an important aspect to consider is that bounding boxes do not accurately segment the object (\ie, objects' shapes are not necessarily rectangular), which may result in background areas that are not part of the object included in the bounding box.
This might result in incorrect associations of irrelevant areas within the object's bounding box with the object.
Therefore, we propose using only a portion of the grid cells that cover the object's area.

More formally, let $(W_o,H_o)\in[0,1]^2\subset\mathbb{R}^2$ be the normalized width and height of an object.
The relative width and height of an object in a specific grid is defined as: ${(W_r,H_r)=(W_o \cdot W_k, H_o \cdot H_k)}$.
In addition, let $(x_{\text{k,center}}, y_{\text{k,center}})$ denote the indices of the cell that the object's center falls in the $k^{\text{th}}$ grid.
We define $p_k$ as the portion of grid cells relative to the grid's size, with regard to the object's center, \ie, the responsible grid cells expand from the object's center to the object's boundaries as a function of $p_k$.
As shown in Figure~\ref{fig:p_obj}, when $p_k=0$, the responsible cells are identical to those of the original YOLO model (single cell), while when $p_k=1$, the responsible cells cover the area of the entire object.

\begin{figure*}[t!]
    \centering
    \includegraphics[width=0.87\linewidth]{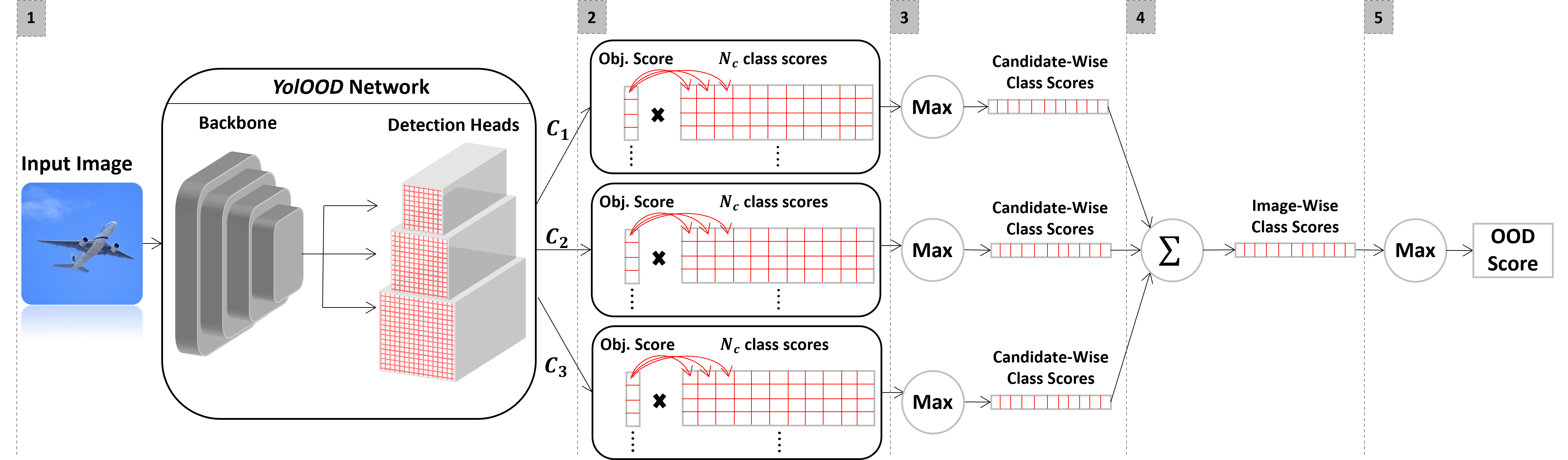}
    \caption{An overview of \emph{YolOOD}'s OOD detection pipeline: 
    (1) the input image is fed to the model, which returns three groups of candidates, each of which corresponds to a different detection head;
    (2) each candidate's objectness score is multiplied by its corresponding class scores; 
    (3) the highest score for each class (candidate-wise class scores) is extracted from each group of candidates; 
    (4) the candidate-wise class scores are aggregated into a single vector of image-wise class scores; and
    (5) the highest score, which will be referred to as the OOD score, is extracted.}
    \label{fig:pipeline}
\end{figure*}

Let $\hat{c}$ denote the corresponding ground-truth value of a predicted candidate $c \in \mathcal{C}_k$.
The objectness score value of $\hat{c}$ at location $(i,j)$ is defined as follows:
\begin{align}
    \begin{split}
        \varphi_u= i \ge x_{\text{k,center}} - p_k \cdot \frac{W_r}{2},\, &
        \varphi_l= j \ge y_{\text{k,center}} - p_k \cdot \frac{H_r}{2} \\
        \varphi_d= i \le x_{\text{k,center}} + p_k \cdot \frac{W_r}{2},\, &
        \varphi_r= j \le y_{\text{k,center}} + p_k \cdot \frac{H_r}{2} \\ 
    \end{split}
\end{align}
\begin{equation}
   \hat{c}_{\text{obj}}(i,j) = 
    \begin{cases}
        1 & \varphi_u \land \varphi_d \land \varphi_l \land \varphi_r \\
        0 & \text{else}
    \end{cases}
\end{equation}

The effect of the different $p_k$ values is discussed in Section~\ref{sec:eval}.

\paragraph{\label{subsec:method:cls_score}Class Scores.}
Each predicted candidate $c \in \mathcal{C}_k$ contains $N_c$ class scores which represent the model's confidence in the presence of each specific class.
Since multiple cells may be responsible for the prediction of an object, a cell can be assigned with multiple classes.
As in the original YOLO training procedure, we train each candidate in a multi-label fashion, \ie, classes are not mutually exclusive.
Formally, the class score for a class $n \in \{1,..,N_c\}$ in $\hat{c}$ at location $(i,j)$ is:
\begin{equation}
    \hat{c}_{\text{cls } n}(i,j) = 
    \begin{cases}
        1 & \text{class } n \text{ is in cell } (i,j)\\
        0 & \text{else}
    \end{cases}
\end{equation}

\subsection{\label{subsec:method:loss}YolOOD Loss Function}

To train YolOOD, we devised a custom loss function that is composed of two components:
\begin{itemize}
    \item Objectness score loss - 
    \begin{equation}
        \mathcal{L}_{\text{obj}} =
        \sum\limits_{c \in \mathcal{C}}
        \mathcal{L}_{\text{BCE}}(c_\text{obj},\hat{c}_\text{obj})
    \end{equation}
    where $\mathcal{L}_{\text{BCE}}$ denotes the binary cross-entropy loss.
    \item Class score loss - 
    \begin{equation}
        \mathcal{L}_{\text{cls}} = 
        \sum\limits_{c \in \mathcal{C}}
        \sum\limits_{n \in \{1,..,N_c\}}
        \hat{c}_{\text{obj}} \cdot \mathcal{L}_{\text{BCE}} (c_\text{cls n},\hat{c}_\text{cls n})
    \end{equation}
\end{itemize}

Finally, the total loss function is:
\begin{equation}
    \mathcal{L}_{\text{total}} = \mathcal{L}_{\text{obj}} + \mathcal{L}_{\text{cls}}
\end{equation}

\subsection{\label{subsec:method:ind}YolOOD as a Multi-label Classifier}

Since the core definition of a multi-label classifier's output is a single vector containing class probabilities, we aggregate YolOOD's output such that the score for each class $n$ is the highest score across all the candidates and is formally defined as:
\begin{equation}
    y_n =
    \max\limits_{c \in \mathcal{C}}
    \{ \sigma(c_{\text{obj}}) \cdot \sigma(c_{\text{cls }n})\}
\end{equation}
where $\sigma$ denotes the sigmoid function.

\subsection{\label{subsec:method:ood}YolOOD for Multi-Label OOD Detection}

Similar to the YOLO candidates' postprocessing (see Section~\ref{subsec:background:OD}), we utilize both the objectness and class scores and propose using \emph{YolOOD} for OOD detection in the following way (visualized in Figure~\ref{fig:pipeline}):
\begin{align}
\label{eq:yolood}
    \text{YolOOD}(x) = 
    \max\limits_{n \in \{1,..,N_c\}}
    \sum\limits_{\mathcal{C}_k \in f_\text{YolOOD}(x)}
    \max\limits_{c \in \mathcal{C}_k}
    \{\sigma(c_\text{obj}) \cdot \sigma(c_{\text{cls } n})\}
\end{align}
\begin{equation}
    G(x,\tau) = 
    \begin{cases}
        1 & \text{YolOOD}(x) \ge \tau \\
        0 & \text{YolOOD}(x) < \tau
    \end{cases}
\end{equation}
where $\tau$ denotes the threshold, which can be selected according to the value that yields a high percentage (\eg, 95\%) of in-distribution data correctly classified by $G(x,\tau)$.

%% file: sections/04_evaluation.tex
\section{\label{sec:eval}Evaluation}

\subsection{\label{subsec:eval:exp_setup}Experimental Setup}

\noindent\textbf{In-distribution datasets.}
We consider the following in-distribution datasets, originally proposed in \cite{hendrycks2019scaling}:
\begin{itemize}[noitemsep]
    \item PASCAL VOC~\cite{everingham2015pascal} - consists of 5,717 training, 5,823 validation, and 10,991 test images across 20 class categories.
    \item MS-COCO~\cite{lin2014microsoft} (2017 version) - consists of 117,266 training, 4,952 validation, and 40,670 test images across 80 class categories.
\end{itemize}
In addition, we propose a new benchmark, which is a subset of the Objects365 dataset~\cite{shao2019objects365}:
\begin{itemize}
    \item $\text{Objects365}_\text{in}$ - a subset of the original dataset which is comprised of the 20 most frequent classes that do not overlap with the classes in the OOD datasets (presented below).
    It consists of 68,723 training, 5,000 validation, and 10,000 test images.
    Further details can be found in the supplementary material.
\end{itemize}
The training and validation images are used in the training process, while the test set is used as an in-distribution set in the OOD detection evaluation.
Furthermore, to demonstrate that our approach is not limited to datasets that contain bounding box annotations, we employ Grounding DINO~\cite{liu2023grounding}, a multi-modal open-set object detection model.
Grounding DINO takes a pair of $\langle\texttt{IMAGE},\texttt{CAPTION}\rangle$ and returns the location of objects in the image based on the provided caption.
In our research, we use Grounding DINO to automatically annotate the training set images, where the caption is simply a concatenation of the category names present in the image (\ie, standard image classification annotations).

\begin{table*}[t!]
\centering
\scalebox{0.68}{
\begin{tabular}{|l|ccc|ccc|}
\hline
\multicolumn{1}{|r|}{$\mathcal{D}_{\text{out}}$} & \multicolumn{3}{c|}{$\text{Objects365}_\text{out}$} &  \multicolumn{3}{c|}{NUS-WIDE} \\
\multicolumn{1}{|r|}{$\mathcal{D}_{\text{in}}$}  & PASCAL-VOC & MS-COCO & $\text{Objects365}_\text{in}$ & PASCAL-VOC & MS-COCO & $\text{Objects365}_\text{in}$ \\
\textbf{Method} & \multicolumn{6}{c|}{\textbf{FPR95 $\downarrow$ / AUROC $\uparrow$ / AUPR $\uparrow$}} \\ \hline\hline
MaxLogit~\cite{hendrycks2019scaling} & 28.91 / 94.96 / 95.32 & 16.39 / 96.90 / 99.17 & 29.95 / 94.33 / 94.38 & 23.60 / 95.99 / 96.05 & 12.16 / 97.53 / 99.24 & 38.07 / 92.62 / 91.48 \\
MSP~\cite{hendrycks2017baseline}     & 50.78 / 88.36 / 88.61 & 46.26 / 86.78 / 95.63 & 65.20 / 83.99 / 84.13 & 47.34 / 89.34 / 88.71 & 40.89 / 88.33 / 95.53 & 78.08 / 78.42 / 76.91 \\
Mahalanobis~\cite{lee2018simple}     & 73.34 / 73.90 / 70.94 & 88.01 / 48.45 / 75.08 & 83.32 / 63.19 / 56.67 & 77.23 / 73.76 / 67.76 & 90.48 / 52.71 / 75.58 & 88.46 / 62.47 / 54.29 \\
ODIN~\cite{liang2018enhancing}       & 28.91 / 94.96 / 95.32 & 16.39 / 96.90 / 99.17 & 29.95 / 94.33 / 94.38 & 23.60 / 95.99 / 96.05 & 12.16 / 97.53 / 99.24 & 38.07 / 92.62 / 91.48 \\
JointEnergy~\cite{wang2021can}       & 27.90 / 95.37 / 96.04 & 14.80 / 97.16 / \textbf{99.28} & 23.13 / 95.84 / \textbf{96.20} & 20.19 / 96.53 / 96.76 & 8.29 / 97.90 / 99.39 & 24.46 / 95.34 / 94.96 \\ \hline
YolOOD-a\footnotemark[1] \cellcolor{Gray}  & 18.37 / 96.10 / 95.85 \cellcolor{Gray} & \cellcolor{Gray}  11.70 / 97.21 / 99.19 & 18.40 / 95.76 / 95.15 \cellcolor{Gray} & 21.24 / 96.29 / 96.08 \cellcolor{Gray} & 7.62 / 98.13 / 99.43  \cellcolor{Gray} &\cellcolor{Gray} 12.19 / 97.64 / 97.29   \\
YolOOD-o\footnotemark[2] \cellcolor{Gray}  & \textbf{16.38} / \textbf{96.60} / \textbf{96.49} \cellcolor{Gray} & \cellcolor{Gray}  \textbf{11.53} / \textbf{97.29} / 99.23 & \textbf{17.24} / \textbf{95.97} / 95.42 \cellcolor{Gray} & \textbf{18.47} / \textbf{96.85} / \textbf{96.77} \cellcolor{Gray} &\textbf{4.40} / \textbf{98.56} / \textbf{99.57}  \cellcolor{Gray} &\cellcolor{Gray}\textbf{9.54} / \textbf{97.99} / \textbf{97.61}   \\
\hline \hline
\end{tabular}}
\caption{Comparison of the OOD detection performance of YolODD vs. state-of-the-art methods.
$\downarrow$ indicates lower values are better, and $\uparrow$ indicates higher values are better. Bold indicates superior results.\\
\scriptsize
\protect\footnotemark[1]{trained using the auto-generated annotations.}
\protect\footnotemark[2]{trained using the original annotations.}}
\label{tab:sota}
\vspace{-0.2cm}
\end{table*}

\noindent\textbf{Out-of-distribution datasets.}
In previous studies \cite{hendrycks2019scaling,wang2021can} proposing solutions for OOD detection in the multi-label setting, the effectiveness of the proposed method was evaluated on a subset of images from the ImageNet-22K~\cite{ILSVRC15} and Textures~\cite{cimpoi2014describing} datasets.
However, these datasets only contain images with a single class category, resulting in an oversimplified setup.
Therefore, we propose two new benchmarks constructed from datasets which contain images associated with multiple class categories and instances, thus reflecting the complexity of the multi-label setting:
\begin{itemize}
    \item $\text{Objects365}_\text{out}$ - a subset of the Objects365 dataset which contains $\sim$200 classes that do not overlap with any of the classes present in the in-distribution datasets (\eg, lamp, tomato), and specifically with the $\text{Objects365}_\text{in}$ subset.
    This subset contains 11,669 images.
    \item NUS-WIDE - a subset of the original NUS-WIDE dataset~\cite{chua2009nus}.
    We remove overlapping classes categories, which leaves us with a subset of 54 categories (\eg, toy, tree, whales).
    This subset contains 13,149 images.
\end{itemize}
Further details about the datasets are presented in the supplementary material.

\paragraph{Metrics.}
In our evaluation, performance is measured with metrics commonly used in the OOD detection domain: (a) \textit{FPR95} - the false positive rate of OOD samples when the true positive rate is at 95\%; (b) \textit{AUROC} - the area under the receiver operating characteristic curve; and (c) \textit{AUPR} - the area under the precision-recall curve.

\noindent\textbf{Networks' architecture.}
We use the latest version of the YOLO object detector, YOLOv5~\cite{yolov5}, pretrained on the MS-COCO dataset.
As explained in Section~\ref{subsec:background:OD}, the network is comprised of a backbone and three detection heads.
YOLOv5 provides several model sizes: nano, small, medium, \etc, each of which contains a different number of learnable parameters for the backbone and detection heads.
We use the YOLOv5 small version (YOLOv5s), which contains $\sim 4.1M$ and $\sim 3M$ learnable parameters in the backbone and detection heads, respectively, which amounts to a total of $\sim 7.1M$ parameters.
To apply our OOD detection approach, we replace the last layer of each detection head with the detection layer described in Section~\ref{subsec:method:det_layer}.

To perform a fair comparison between the proposed method and other state-of-the-art OOD detection methods, we train a multi-label image classifier based on the backbone of YOLOv5s.
To this end, we replace the detection heads with three fully connected layers, resulting in a similar-sized network (total of $\sim 7.1M$ parameters).
This network is referred to as \textit{YOLO-cls} in the evaluation.

\noindent\textbf{Training details.}
For each in-distribution dataset, we fine-tune a pair of \textit{YolOOD} and \textit{YOLO-cls} models using the backbone's pretrained weights.
More precisely, we fine-tune five pairs, each initialized with a different seed.
The results presented in the paper are averaged across them (the complete results, including the standard deviation values, are included in the supplementary material).
We use the Adam optimizer~\cite{kingma2014adam} with an initial learning rate of $10^{-5}$ and $10^{-4}$ respectively for the backbone and the remaining layers.
The learning rate is reduced by a factor of 10 if the mAP on the validation set does not improve for two consecutive epochs.
For the \textit{YOLO-cls} model, we apply the logistic sigmoid function on the outputs of the classification layer (\ie, logits) for multi-label training.
For both of the models, the images are resized to $640\times 640$ pixels and applied with color-based augmentations and geometric transformations.
The mAP's for YolOOD are on par with YOLO-cls on all of the in-distribution datasets evaluated ($\sim$1-2\% difference).
Detailed results are presented in the supplementary material.

\subsection{\label{subsec:eval:results}Results}
\begin{figure}[t]
    \centering
    \includegraphics[width=0.7\linewidth]{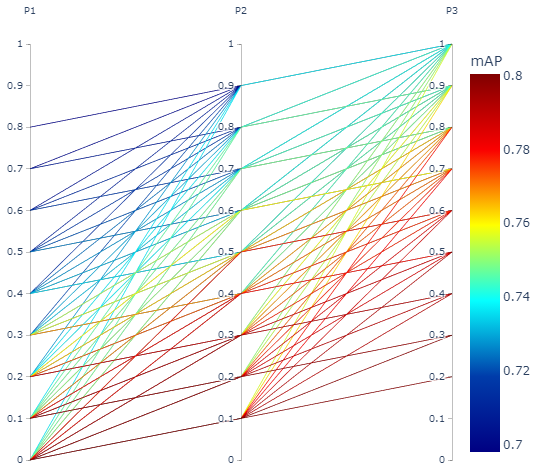}
    \caption{Models' mAP for different $p_k$ combinations on the COCO dataset.}
    \label{fig:pk_coco}
\end{figure}

\paragraph{Effect of responsible grid cell percentage $\boldsymbol{p_k}$.}
We characterize the effect of the percentage ${\{p_k \vert k \in \{1,2,3\}\}}$ of responsible cells (described in Section~\ref{subsec:method:obj_score}), where $p_1$ (resp. $p_3$) represents the smallest (resp. largest) detection head.
We perform an extensive evaluation to determine the effect of different $p_k$ combinations, where $p_k$ is selected from 11 evenly spaced numbers in the range $[0,1]$.
To limit the number of possible combinations, we set a constraint such that $p_3 > p_2 > p_1$, based on the fact that the grid's resolution increases in each subsequent detection head, resulting in 165 different combinations.
An illustration of the results obtained on the COCO dataset is provided in Figure~\ref{fig:pk_coco}. 
After training, we sort all of the models according to their in-distribution mAP and select the 20 best-performing ones.
Then, we count the number of occurrences of each $p_k$ value (each $p_k$ is counted independently, not as a triplet), and select the most frequent values.
After aggregating the results over all in-distribution datasets, we found that the best configuration is $(p_1,p_2,p_3)=(0.0,0.1,0.5)$.
As expected, the detection head with the highest resolution ($k=3$) benefits greatly from a large portion of responsible cells, while the detection head with the smallest resolution works best with just a single responsible cell.
It should be emphasized that we recommend using this configuration for all datasets, \ie, $p_k$ is not a hyperparameter that should be tuned.
The complete results can be found in the supplementary material.

\paragraph{YolOOD vs. state-of-the-art OOD detection methods.}
We compare our approach to state-of-the-art OOD detection methods using the \emph{YOLO-cls} network: 
(a) MaxLogit~\cite{hendrycks2019scaling}, 
(b) Maximum Softmax Probablility (MSP)~\cite{hendrycks2017baseline},
(c) ODIN~\cite{liang2018enhancing}, 
(d) Mahalanobis~\cite{lee2018simple}, and 
(e) JointEnergy~\cite{wang2021can}.
In Table~\ref{tab:sota} we can see that YolOOD outperforms all baselines and state-of-the-art approaches, including the multi-label OOD detection method, JointEnergy.
Specifically, when the networks are trained on the PASCAL-VOC, MS-COCO, and $\text{Objects365}_\text{in}$ datasets with the \emph{auto-generated} annotations (\ie, constructed using Grouding DINO with solely standard classification annotations), and the OOD detection is evaluated on the $\text{Objects365}_\text{out}$ OOD dataset, YolOOD reduces the FPR95 by 9.53\%, 3.10\%, and 4.73\%, respectively, compared to the best-performing method.

It should also be noted that the Mahalanobis method shows poor performance compared to the other methods examined.
In ~\cite{wang2021can}, the authors hypothesized that the Mahalanobis method may not be well suited for the multi-label task, since it is based on the assumption that feature representation forms class-conditional Gaussian distributions.
This assumption also holds in our case, however from a different perspective - the \emph{YOLO-cls} backbone used for the evaluation was originally trained for the object detection task and therefore learned different feature representations.
Furthermore, the performance of ODIN and MaxLogit is similar , since the best hyperparameter configuration found for ODIN is equivalent to MaxLogit (a special case where the temperature is set at one and the magnitude of noise is set at zero), which correlates with the results in \cite{wang2021can}.

\paragraph{\label{par:obj_cls}Objectness score vs. class score vs. joint score.}
We also perform an analysis to examine the effect of different aggregation methods on the candidates' output scores, \ie, objectness and class scores.
We consider two different approaches in addition to the regular YolOOD approach described in Equation~\ref{eq:yolood} (which uses a joint probability between the objectness and class scores):
\begin{align}
    & \text{YolOOD}_{\text{Cls}}(x) =
    \max\limits_{n \in \{1,..,N_c\}}
    \sum\limits_{\mathcal{C}_k \in f_\text{YolOOD}(x)}
    \max\limits_{c \in \mathcal{C}_k}
    \{\sigma(c_{\text{cls } n})\} \\
    & \text{YolOOD}_{\text{Obj}}(x) =
    \sum\limits_{\mathcal{C}_k \in f_\text{YolOOD}(x)}
    \max\limits_{c \in \mathcal{C}_k}
    \{\sigma(c_{obj})\} 
\end{align}

\begin{figure}[t]
    \centering
    \includegraphics[width=1.0\linewidth]{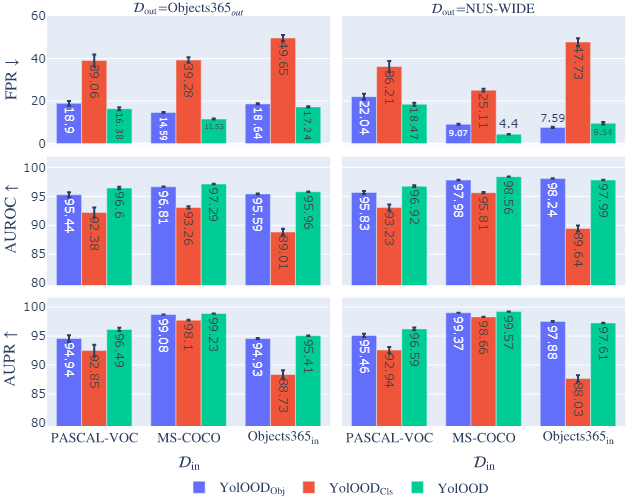}
    \caption{A comparison of the impact of different aggregation methods on YolOOD candidates' output scores (described in Section~\ref{par:obj_cls}).}
    \label{fig:obj_cls}
\end{figure}

In Figure~\ref{fig:obj_cls}, we can see that when using the objectness and class scores separately ($\text{YolOOD}_{\text{Obj}}$ and $\text{YolOOD}_{\text{Cls}}$, respectively), the objectness score contributes more to the ability to distinguish between in-distribution and OOD samples.
Nonetheless, when using a joint probability (\ie, a combination of the scores), the OOD detection performance exceeds the performance of both individually.

\paragraph{Effectiveness of different aggregation functions.}
Since YolOOD's output is comprised of candidates from three detection heads, we consider the use of different aggregation functions between them, \ie, instead of combining all of the candidates from all three detection heads into one large set $\mathcal{C}=\bigcup_{k}\mathcal{C}_k$, we extract the best candidate from each set $\mathcal{C}_k$ and only then apply the aggregation function.
In addition, we also examine the effect of different aggregation functions on the class score output vector (similar to ~\cite{wang2021can} which proposed summing energies over all of the labels).
More formally, we replace Equation~\ref{eq:yolood} with the following:
\begin{equation}
    \agga\limits_{n \in \{1,..,N_c\}}
    \aggb\limits_{\mathcal{C}_k \in f_\text{YolOOD}(x)}
    \max\limits_{c \in \mathcal{C}_k}
    \{\sigma(c_\text{obj}) \cdot \sigma(c_{\text{cls } n})\}
\end{equation}
where $\agga$ can either be the summation or max function, and $\aggb$ can be one of the following functions: summation, multiplication, or max.

In general, the results presented in Table~\ref{tab:agg_func} show that on most in-distribution datasets, the approach that yields the best results is summing the scores over the different candidate sets and then extracting the maximum value of all of the class scores.
It is interesting to see that when summing over the class scores (Table~\ref{tab:agg_func} bottom), multiplication between the candidate sets works better than the sum or max functions.

\begin{table}[t]
\centering
\scalebox{0.61}{
\begin{tabular}{clccc}
\hline
 & \multicolumn{1}{r}{$\mathcal{D}_{\text{in}}$}  & PASCAL-VOC &  MS-COCO & $\text{Objects365}_\text{in}$\\
Class Agg. & Head Agg. & \multicolumn{3}{c}{\textbf{FPR95 $\downarrow$ / AUROC $\uparrow$ / AUPR $\uparrow$}} \\ 
\hline\hline
\multirow{3}{*}{Max}
& Max      & 24.10 / 95.63 / 95.61 & \,  8.55 / \textbf{98.01} / \textbf{99.43} & 21.01 / 95.70 / 95.08 \\
& Multiply & 19.76 / 96.02 / 95.74 & 10.55 / 97.59 / 99.30 & 18.22 / 96.23 / 95.68 \\
& Sum      & \textbf{17.43} / \textbf{96.73} / \textbf{96.63} & \, \textbf{7.97} / 97.93 / 99.41 & \textbf{13.39} / \textbf{96.98} / \textbf{96.52} \\
\hline
\multirow{3}{*}{Sum}
& Max      & 44.56 / 78.31 / 70.89 & 54.70 / 80.59 / 92.21 & 35.26 / 85.67 / 79.28 \\
& Multiply & \textbf{19.58} / \textbf{96.04} / \textbf{95.80} & \, \textbf{9.61} / \textbf{97.81} / \textbf{99.38} & \textbf{17.55} / \textbf{96.39} / \textbf{95.94} \\
& Sum      & 40.11 / 84.91 / 81.83 & 39.50 / 90.20 / 96.71 & 26.18 / 93.68 / 92.39 \\
\hline \hline
\end{tabular}}
\caption{OOD detection performance when using different combinations of aggregation functions for YolOOD's detection heads and class scores output vector.
The results are averaged across the OOD datasets.}
\label{tab:agg_func}
\end{table}

\paragraph{Combining YolOOD with JointEnergy.}
Wang~\etal~\cite{wang2021can} presented the JointEnergy technique in the following way:
\begin{equation}
    \begin{gathered}
    E_{y_n}(x) = -\log (1 + e^{f_{y_n}(x)}) \\ 
    E_{\text{joint}}(x) = \sum\nolimits_{n=1}^{N_c} -E_{y_n}(x)
    \end{gathered}
\end{equation}
where $f_{y_n}(x)$ denotes the logit of the $\text{n}^\text{th}$ class.

Since YolOOD's output can be transformed into a single vector of class probabilities (as shown in Section~\ref{subsec:method:ind}), JointEnergy can be applied and used with the YolOOD architecture.
Therefore, instead of applying the sigmoid function to the objectness and class scores (Equation~\ref{eq:yolood}), we apply the energy function $E_{y_n}$ and combine label-wise energies over all labels for a single OOD score.
Formally, the combination of YolOOD and JointEnergy can be written as follows:
\begin{multline*}
    \text{YolOOD}_{\text{JointEnergy}}(x)= \\
    \sum\limits_{n \in \{1,..,N_c\}}
    \sum\limits_{\mathcal{C}_k \in f_\text{YolOOD}(x)}
    -\max\limits_{c \in \mathcal{C}_k}
    \{E_{y_n}(c_\text{obj}) \cdot E_{y_n}(c_{\text{cls } n})\}
\end{multline*}

We compare the results obtained with $\text{YolOOD}_{\text{JointEnergy}}$ to the performance of JointEnergy when applied to the YOLO-cls network.
We observe that $\text{YolOOD}_{\text{JointEnergy}}$ outperforms JointEnergy when applied to the YOLO-cls network on most in-distribution and OOD datasets across all metrics.
For example, when using $\text{Objects365}_\text{out}$ as the OOD dataset, the FPR95 metric improves by 10.49\%, 7.7\%, and 7.22\% when the networks are trained on the PASCAL-VOC, MS-COCO, and $\text{Objects365}_\text{in}$, respectively.
Moreover, in some cases, the performance of $\text{YolOOD}_{\text{JointEnergy}}$ even exceeds the performance obtained using the regular YolOOD score (\eg, when MS-COCO and $\text{Objects365}_\text{out}$ are used as the in-distribution and OOD datasets, respectively, the FPR95 decreases from 11.53\% to 7.1\%).

\paragraph{YolOOD vs. a vanilla YOLO}
We also examine the advantage of using YolOOD over a vanilla YOLO (\ie, a standard object detector).
We train a YOLO detector using the same configuration used to train YolOOD for a fair comparison: pretrained weights are only used in the backbone layers and no custom augmentations are used (\eg, mosaic~\cite{bochkovskiy2020yolov4}).
The OOD score functions for the vanilla YOLO model are similar to those of YolOOD, differing only at the anchor boxes level, which results in $3\times$ more candidates.

\begin{figure}[t!]
    \centering
    \begin{subfigure}{0.47\linewidth}
        \includegraphics[width=0.85\linewidth]{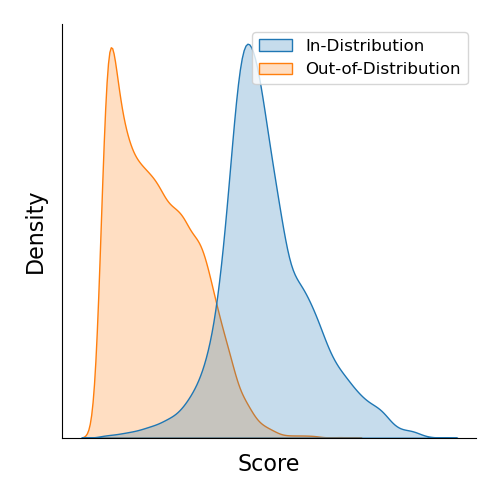}
        \caption{YolOOD (FPR95=16.38\%)}
        \label{subfig:yolood_vs_yolo:yolood}
    \end{subfigure}
    \begin{subfigure}{0.47\linewidth}
        \includegraphics[width=0.85\linewidth]{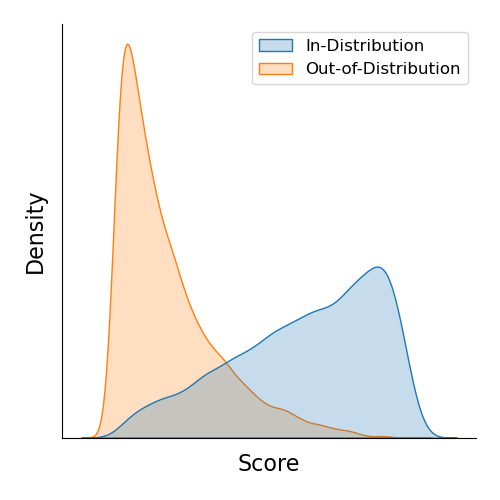}
        \caption{YOLO (FPR95=40.57\%)}
        \label{subfig:yolood_vs_yolo:yolo}
    \end{subfigure}
    \caption{Score distribution when using PASCAL-VOC as the in-distribution dataset and $\text{Objects365}_\text{out}$ as the OOD dataset.}
    \label{fig:yolood_vs_yolo}
\end{figure}

Our evaluation demonstrates that YolOOD outperforms vanilla YOLO on all of the in-distribution and OOD datasets examined.
An example of this is provided in Figure~\ref{fig:yolood_vs_yolo}, which presents the distribution of YolOOD and vanilla YOLO scores when using PASCAL-VOC and $\text{Objects365}_\text{out}$ as in-distribution and OOD datasets, respectively.
As can be seen, there are notable distinctions between the two models. 
Vanilla YOLO (Figure~\ref{subfig:yolood_vs_yolo:yolo}) excels at assigning low scores to OOD data but faces challenges in definitively identifying in-distribution data, as evidenced by the uniformly distributed scores obtained.
On the other hand, YolOOD (Figure~\ref{subfig:yolood_vs_yolo:yolood}) performs well on OOD data, while substantially enhancing in-distribution detection.
We hypothesize that the difference in the models' performance is mainly the result of the fact that (a) the loss function in the vanilla YOLO largely focuses on improving the bounding box coordinates' regression, and (b) YolOOD uses more responsible grid cells for the detection of each object, capturing broader representations.

%% file: sections/05_related_work.tex
\section{\label{sec:related}Related Work}
\subsection{\label{subsec:related:od}Object Detection}

Object detectors have been studied extensively over the last few years, demonstrating state-of-the-art performance, and various solutions aimed at identifying the objects present in an image and their precise location (\ie, bounding box) have been proposed.
Modern deep learning-based object detectors are usually composed of two components, a backbone network and a head, which is used to predict the bounding boxes and classes of existing objects in an image.
Broadly, there are two types of models: one-stage detectors (\eg, SSD~\cite{liu2016ssd} and YOLO~\cite{redmon2018yolov3,bochkovskiy2020yolov4,yolov5}) and two-stage detectors (\eg, Mask R-CNN~\cite{he2017mask} and Faster R-CNN~\cite{ren2015faster}).
Object detectors developed in recent years often insert additional layers between the backbone and the head, which are used to collect feature maps from different stages~\cite{lin2017feature,tan2020efficientdet}.

\subsection{\label{subsec:related:ood}Out-of-Distribution Detection}

Deep neural networks' overconfidence for OOD data was first noted by \cite{nguyen2015deep}.
Several baseline approaches have been proposed to tackle this problem; for example, Hendrycks~\etal proposed two baselines: Maximum Softmax Probability (MSP)~\cite{hendrycks2017baseline} and MaxLogit, which uses the highest score from the classifier's last layer as an OOD score~\cite{hendrycks2019scaling}.
In recent years, OOD detection has attracted the attention of the machine learning research community, and researchers have proposed methods aimed at improving OOD uncertainty estimation, such as: 
(1) ODIN~\cite{liang2018enhancing}, which combines input preprocessing and temperature scaling; 
(2) the Mahalanobis~\cite{lee2018simple} distance-based approach, which utilizes the network's internal feature representations; 
(3) the gradient-based GradNorm~\cite{huang2021importance} score; 
(4) ReAct~\cite{sun2021react}, which rectifies activation values; 
(5) IsoMax~\cite{macedo2021entropic}, which proposes isotropy maximization loss; and 
(6) the energy score~\cite{liu2020energy}.
These studies only addressed the multi-class classification task, and the topic of OOD detection in the multi-label domain remains largely underexplored.
OOD detection in the multi-label domain has only been addressed by \cite{hendrycks2019scaling} who proposed the MaxLogit baseline, and \cite{wang2021can}, who proposed JointEnergy which combines label-wise energies over all labels.

It should be noted that the OOD detection methods proposed in many prior studies require \emph{external} OOD data (in addition to the OOD test dataset) to improve the detector's robustness.
In some cases, the model is provided with OOD samples for hyperparameter tuning~\cite{liang2018enhancing}, while in other cases, these samples are used for negative learning in which the model is explicitly trained on images that do not contain in-distribution data~\cite{hendrycks2018deep,mohseni2020self}.
In contrast, object detection models in general, and YOLO in particular, do not require any type of external OOD data to model irrelevant objects and background areas.

%% file: sections/06_conclusion.tex
\section{\label{sec:conclusion}Conclusion}

In this paper, we presented YolOOD -- an OOD detection approach for the underexplored multi-label classification domain which utilizes the main concepts of object detectors.
We demonstrated how all parts of an input image can be exploited to model both in-distribution and OOD data, without depending on external data sources.
In our evaluation, we performed a comprehensive set of experiments on various benchmark datasets and presented new benchmark datasets that better reflect the complexity of the multi-label domain.
Our approach achieved state-of-the-art performance compared to the examined OOD detection methods and a standard YOLO object detector.
In future work, we would like to explore the possibility of modeling an image at a deeper granularity level (instead of bounding boxes), \eg, using pixel-wise annotations.

%% file: main.bbl
\begin{thebibliography}{34}
\providecommand{\natexlab}[1]{#1}
\providecommand{\url}[1]{\texttt{#1}}
\expandafter\ifx\csname urlstyle\endcsname\relax
  \providecommand{\doi}[1]{doi: #1}\else
  \providecommand{\doi}{doi: \begingroup \urlstyle{rm}\Url}\fi

\bibitem[Amit et~al.(2021)Amit, Levy, Rosenberg, Shabtai, and Elovici]{amit2021food}
Guy Amit, Moshe Levy, Ishai Rosenberg, Asaf Shabtai, and Yuval Elovici.
\newblock Food: Fast out-of-distribution detector.
\newblock In \emph{2021 International Joint Conference on Neural Networks (IJCNN)}, pages 1--8. IEEE, 2021.

\bibitem[Bochkovskiy et~al.(2020)Bochkovskiy, Wang, and Liao]{bochkovskiy2020yolov4}
Alexey Bochkovskiy, Chien-Yao Wang, and Hong-Yuan~Mark Liao.
\newblock Yolov4: Optimal speed and accuracy of object detection.
\newblock \emph{arXiv preprint arXiv:2004.10934}, 2020.

\bibitem[Chen et~al.(2019)Chen, Pang, Wang, Xiong, Li, Sun, Feng, Liu, Shi, Ouyang, et~al.]{chen2019hybrid}
Kai Chen, Jiangmiao Pang, Jiaqi Wang, Yu Xiong, Xiaoxiao Li, Shuyang Sun, Wansen Feng, Ziwei Liu, Jianping Shi, Wanli Ouyang, et~al.
\newblock Hybrid task cascade for instance segmentation.
\newblock In \emph{Proceedings of the IEEE conference on computer vision and pattern recognition}, pages 4974--4983, 2019.

\bibitem[Chen et~al.(2014)Chen, Papandreou, Kokkinos, Murphy, and Yuille]{chen2014semantic}
Liang-Chieh Chen, George Papandreou, Iasonas Kokkinos, Kevin Murphy, and Alan~L Yuille.
\newblock Semantic image segmentation with deep convolutional nets and fully connected crfs.
\newblock \emph{arXiv preprint arXiv:1412.7062}, 2014.

\bibitem[Chua et~al.(2009)Chua, Tang, Hong, Li, Luo, and Zheng]{chua2009nus}
Tat-Seng Chua, Jinhui Tang, Richang Hong, Haojie Li, Zhiping Luo, and Yantao Zheng.
\newblock Nus-wide: a real-world web image database from national university of singapore.
\newblock In \emph{Proceedings of the ACM international conference on image and video retrieval}, pages 1--9, 2009.

\bibitem[Cimpoi et~al.(2014)Cimpoi, Maji, Kokkinos, Mohamed, and Vedaldi]{cimpoi2014describing}
Mircea Cimpoi, Subhransu Maji, Iasonas Kokkinos, Sammy Mohamed, and Andrea Vedaldi.
\newblock Describing textures in the wild.
\newblock In \emph{Proceedings of the IEEE conference on computer vision and pattern recognition}, pages 3606--3613, 2014.

\bibitem[Everingham et~al.(2015)Everingham, Eslami, Van~Gool, Williams, Winn, and Zisserman]{everingham2015pascal}
Mark Everingham, SM Eslami, Luc Van~Gool, Christopher~KI Williams, John Winn, and Andrew Zisserman.
\newblock The pascal visual object classes challenge: A retrospective.
\newblock \emph{International journal of computer vision}, 111\penalty0 (1):\penalty0 98--136, 2015.

\bibitem[He et~al.(2015)He, Zhang, Ren, and Sun]{he2015delving}
Kaiming He, Xiangyu Zhang, Shaoqing Ren, and Jian Sun.
\newblock Delving deep into rectifiers: Surpassing human-level performance on imagenet classification.
\newblock In \emph{Proceedings of the IEEE international conference on computer vision}, pages 1026--1034, 2015.

\bibitem[He et~al.(2017)He, Gkioxari, Doll{\'a}r, and Girshick]{he2017mask}
Kaiming He, Georgia Gkioxari, Piotr Doll{\'a}r, and Ross Girshick.
\newblock Mask r-cnn.
\newblock In \emph{Proceedings of the IEEE international conference on computer vision}, pages 2961--2969, 2017.

\bibitem[Hendrycks and Gimpel(2017)]{hendrycks2017baseline}
Dan Hendrycks and Kevin Gimpel.
\newblock A baseline for detecting misclassified and out-of-distribution examples in neural networks.
\newblock In \emph{International Conference on Learning Representations}, 2017.

\bibitem[Hendrycks et~al.(2019{\natexlab{a}})Hendrycks, Basart, Mazeika, Mostajabi, Steinhardt, and Song]{hendrycks2019scaling}
Dan Hendrycks, Steven Basart, Mantas Mazeika, Mohammadreza Mostajabi, Jacob Steinhardt, and Dawn Song.
\newblock Scaling out-of-distribution detection for real-world settings.
\newblock \emph{arXiv preprint arXiv:1911.11132}, 2019{\natexlab{a}}.

\bibitem[Hendrycks et~al.(2019{\natexlab{b}})Hendrycks, Mazeika, and Dietterich]{hendrycks2018deep}
Dan Hendrycks, Mantas Mazeika, and Thomas Dietterich.
\newblock Deep anomaly detection with outlier exposure.
\newblock In \emph{International Conference on Learning Representations}, 2019{\natexlab{b}}.

\bibitem[Huang et~al.(2021)Huang, Geng, and Li]{huang2021importance}
Rui Huang, Andrew Geng, and Yixuan Li.
\newblock On the importance of gradients for detecting distributional shifts in the wild.
\newblock \emph{Advances in Neural Information Processing Systems}, 34:\penalty0 677--689, 2021.

\bibitem[Jocher(2021)]{yolov5}
Glenn Jocher.
\newblock ultralytics/yolov5: v6.0 - yolov5n 'nano' models, roboflow integration, tensorflow export, opencv dnn support, 2021.

\bibitem[Kingma and Ba(2014)]{kingma2014adam}
Diederik~P Kingma and Jimmy Ba.
\newblock Adam: A method for stochastic optimization.
\newblock \emph{arXiv preprint arXiv:1412.6980}, 2014.

\bibitem[Krizhevsky et~al.(2012)Krizhevsky, Sutskever, and Hinton]{krizhevsky2012imagenet}
Alex Krizhevsky, Ilya Sutskever, and Geoffrey~E Hinton.
\newblock Imagenet classification with deep convolutional neural networks.
\newblock In \emph{Advances in neural information processing systems}, pages 1097--1105, 2012.

\bibitem[Lee et~al.(2018)Lee, Lee, Lee, and Shin]{lee2018simple}
Kimin Lee, Kibok Lee, Honglak Lee, and Jinwoo Shin.
\newblock A simple unified framework for detecting out-of-distribution samples and adversarial attacks.
\newblock \emph{Advances in neural information processing systems}, 31, 2018.

\bibitem[Liang et~al.(2018)Liang, Li, and Srikant]{liang2018enhancing}
Shiyu Liang, Yixuan Li, and R Srikant.
\newblock Enhancing the reliability of out-of-distribution image detection in neural networks.
\newblock In \emph{6th International Conference on Learning Representations, ICLR 2018}, 2018.

\bibitem[Lin et~al.(2014)Lin, Maire, Belongie, Hays, Perona, Ramanan, Doll{\'a}r, and Zitnick]{lin2014microsoft}
Tsung-Yi Lin, Michael Maire, Serge Belongie, James Hays, Pietro Perona, Deva Ramanan, Piotr Doll{\'a}r, and C~Lawrence Zitnick.
\newblock Microsoft coco: Common objects in context.
\newblock In \emph{European conference on computer vision}, pages 740--755. Springer, 2014.

\bibitem[Lin et~al.(2017)Lin, Doll{\'a}r, Girshick, He, Hariharan, and Belongie]{lin2017feature}
Tsung-Yi Lin, Piotr Doll{\'a}r, Ross Girshick, Kaiming He, Bharath Hariharan, and Serge Belongie.
\newblock Feature pyramid networks for object detection.
\newblock In \emph{Proceedings of the IEEE conference on computer vision and pattern recognition}, pages 2117--2125, 2017.

\bibitem[Liu et~al.(2023)Liu, Zeng, Ren, Li, Zhang, Yang, Li, Yang, Su, Zhu, et~al.]{liu2023grounding}
Shilong Liu, Zhaoyang Zeng, Tianhe Ren, Feng Li, Hao Zhang, Jie Yang, Chunyuan Li, Jianwei Yang, Hang Su, Jun Zhu, et~al.
\newblock Grounding dino: Marrying dino with grounded pre-training for open-set object detection.
\newblock \emph{arXiv preprint arXiv:2303.05499}, 2023.

\bibitem[Liu et~al.(2016)Liu, Anguelov, Erhan, Szegedy, Reed, Fu, and Berg]{liu2016ssd}
Wei Liu, Dragomir Anguelov, Dumitru Erhan, Christian Szegedy, Scott Reed, Cheng-Yang Fu, and Alexander~C Berg.
\newblock Ssd: Single shot multibox detector.
\newblock In \emph{European conference on computer vision}, pages 21--37. Springer, 2016.

\bibitem[Liu et~al.(2020)Liu, Wang, Owens, and Li]{liu2020energy}
Weitang Liu, Xiaoyun Wang, John Owens, and Yixuan Li.
\newblock Energy-based out-of-distribution detection.
\newblock \emph{Advances in Neural Information Processing Systems}, 33:\penalty0 21464--21475, 2020.

\bibitem[Mac{\^e}do et~al.(2021)Mac{\^e}do, Ren, Zanchettin, Oliveira, and Ludermir]{macedo2021entropic}
David Mac{\^e}do, Tsang~Ing Ren, Cleber Zanchettin, Adriano~LI Oliveira, and Teresa Ludermir.
\newblock Entropic out-of-distribution detection.
\newblock In \emph{2021 International Joint Conference on Neural Networks (IJCNN)}, pages 1--8. IEEE, 2021.

\bibitem[Mohseni et~al.(2020)Mohseni, Pitale, Yadawa, and Wang]{mohseni2020self}
Sina Mohseni, Mandar Pitale, JBS Yadawa, and Zhangyang Wang.
\newblock Self-supervised learning for generalizable out-of-distribution detection.
\newblock In \emph{Proceedings of the AAAI Conference on Artificial Intelligence}, pages 5216--5223, 2020.

\bibitem[Nguyen et~al.(2015)Nguyen, Yosinski, and Clune]{nguyen2015deep}
Anh Nguyen, Jason Yosinski, and Jeff Clune.
\newblock Deep neural networks are easily fooled: High confidence predictions for unrecognizable images.
\newblock In \emph{Proceedings of the IEEE conference on computer vision and pattern recognition}, pages 427--436, 2015.

\bibitem[Redmon and Farhadi(2018)]{redmon2018yolov3}
Joseph Redmon and Ali Farhadi.
\newblock Yolov3: An incremental improvement.
\newblock \emph{arXiv preprint arXiv:1804.02767}, 2018.

\bibitem[Redmon et~al.(2016)Redmon, Divvala, Girshick, and Farhadi]{redmon2016you}
Joseph Redmon, Santosh Divvala, Ross Girshick, and Ali Farhadi.
\newblock You only look once: Unified, real-time object detection.
\newblock In \emph{Proceedings of the IEEE conference on computer vision and pattern recognition}, pages 779--788, 2016.

\bibitem[Ren et~al.(2015)Ren, He, Girshick, and Sun]{ren2015faster}
Shaoqing Ren, Kaiming He, Ross Girshick, and Jian Sun.
\newblock Faster r-cnn: Towards real-time object detection with region proposal networks.
\newblock \emph{Advances in neural information processing systems}, 28, 2015.

\bibitem[Russakovsky et~al.(2015)Russakovsky, Deng, Su, Krause, Satheesh, Ma, Huang, Karpathy, Khosla, Bernstein, Berg, and Fei-Fei]{ILSVRC15}
Olga Russakovsky, Jia Deng, Hao Su, Jonathan Krause, Sanjeev Satheesh, Sean Ma, Zhiheng Huang, Andrej Karpathy, Aditya Khosla, Michael Bernstein, Alexander~C. Berg, and Li Fei-Fei.
\newblock {ImageNet Large Scale Visual Recognition Challenge}.
\newblock \emph{International Journal of Computer Vision (IJCV)}, 115\penalty0 (3):\penalty0 211--252, 2015.

\bibitem[Shao et~al.(2019)Shao, Li, Zhang, Peng, Yu, Zhang, Li, and Sun]{shao2019objects365}
Shuai Shao, Zeming Li, Tianyuan Zhang, Chao Peng, Gang Yu, Xiangyu Zhang, Jing Li, and Jian Sun.
\newblock Objects365: A large-scale, high-quality dataset for object detection.
\newblock In \emph{Proceedings of the IEEE/CVF international conference on computer vision}, pages 8430--8439, 2019.

\bibitem[Sun et~al.(2021)Sun, Guo, and Li]{sun2021react}
Yiyou Sun, Chuan Guo, and Yixuan Li.
\newblock React: Out-of-distribution detection with rectified activations.
\newblock \emph{Advances in Neural Information Processing Systems}, 34:\penalty0 144--157, 2021.

\bibitem[Tan et~al.(2020)Tan, Pang, and Le]{tan2020efficientdet}
Mingxing Tan, Ruoming Pang, and Quoc~V Le.
\newblock Efficientdet: Scalable and efficient object detection.
\newblock In \emph{Proceedings of the IEEE/CVF conference on computer vision and pattern recognition}, pages 10781--10790, 2020.

\bibitem[Wang et~al.(2021)Wang, Liu, Bocchieri, and Li]{wang2021can}
Haoran Wang, Weitang Liu, Alex Bocchieri, and Yixuan Li.
\newblock Can multi-label classification networks know what they don’t know?
\newblock \emph{Advances in Neural Information Processing Systems}, 34:\penalty0 29074--29087, 2021.

\end{thebibliography}
